\documentclass[a4paper,fleqn]{cas-sc}

\usepackage[authoryear]{natbib}

\usepackage{tabularx}
\usepackage{multirow}
\usepackage{makecell}
\usepackage{placeins}
\ExplSyntaxOn
\NewDocumentEnvironment{fixedfigure}{}
  {
    \par\addvspace{\intextsep}
    \noindent\begin{minipage}{\linewidth}
    \__reset_fig:
    \dim_set:Nn \l_fig_width_dim {\linewidth}
    \centering\sffamily\small
    \cs_set:cpn {@captype} {figure}
    \cs_set_eq:cN {@makecaption} \__make_fig_caption:nn
  }
  {
    \end{minipage}
    \par\addvspace{\intextsep}
  }
\cs_set:Npn \__first_footerline: {}

\cs_set:Npn \__cas_head: {}
\use:c {ps@cas}
\ExplSyntaxOff

\begin{document}
\let\WriteBookmarks\relax
\def\floatpagepagefraction{1}
\def\textpagefraction{.001}

\title[mode = title]{Same Predictions, Different Reasons: The Effect of
Quantization on Model Explanations}

\author[1]{Kazi Kamruzzaman Rabbi}

\ead{kazi.kamruzzaman@bracu.ac.bd}

\credit{Conceptualization, Methodology, Software, Writing - Original draft preparation}

\affiliation[1]{organization={BRAC University},
                city={Dhaka},
                country={Bangladesh}}

\author[2]{Md. Zami Al Zunaed Farabe}
\ead{md.zunaed.farabe@gmail.com}

\affiliation[2]{organization={Department of Computer Science and Engineering, Bangladesh University of Engineering and Technology (BUET)},
                city={Dhaka},
                country={Bangladesh}}

\author[2]{M. Sohel Rahman}
\ead{msrahman@cse.buet.ac.bd}


\begin{abstract}
Post-training quantization (PTQ) has become a practical solution for deploying deep learning models on resource-constrained edge devices by compressing high-precision floating-point weights into low-precision representations without requiring retraining. Past research has demonstrated that quantization largely preserves classification accuracy; however, whether it also preserves the model's internal reasoning remains an open question. This study presents a systematic evaluation on how static PTQ affects the interpretability / explainability of five widely used CNN architectures: VGG19, ResNet18, EfficientNet-B0, DenseNet161, and MobileNetV2 at INT8 and INT4 precision. We employ a dual interpretability framework that combines Grad-CAM for spatial attention analysis with LIME for input-level feature attribution, and systematically compare full-precision and quantized models on two binary classification datasets. Interpretability is evaluated using three complementary metrics: the Pearson correlation coefficient, structural similarity index, and top-20\% IoU to capture distributional and structural variations in model explanations, supplemented by deletion/insertion faithfulness analysis. The results show that classification accuracy is not a reliable indicator of interpretability stability under reduced precision. DenseNet161 maintains strong feature consistency across both precision levels, whereas EfficientNet-B0, despite achieving competitive spatial attention and classification accuracy at INT8 precision, exhibits a substantial degradation in input-level feature attribution. These findings have direct implications for the trustworthy deployment of quantized models in applications with high interpretability requirements, demonstrating that architecture selection is as important as the quantization strategy.
\end{abstract}


\maketitle

\section{Introduction}

Deep learning eliminates the need for manual feature engineering by
enabling models to autonomously learn features from large datasets using
architectures with millions of parameters
\citep{krizhevsky2012imagenet,lecun2015deep}. Different deep learning
techniques provide strong performance in tasks such as object detection,
segmentation, and classification. However, these computationally heavy
models face deployment challenges in resource-limited edge devices such
as embedded systems, IoT devices, and smartphones \citep{chen2020deep}.
Quantization addresses this by converting high-precision floating-point
weights into lower-precision formats, reducing model size, power use, and
inference time
\citep{gholami2022survey,rokh2023comprehensive,wei2024advances}.

Two primary quantization approaches exist, namely, Post-Training
Quantization (PTQ), which avoids the significant computational cost and
complexity of retraining but can cause accuracy loss, and
Quantization-Aware Training (QAT), which integrates quantization during
training, requires full retraining, and is resource-intensive, but
preserves near-original accuracy. Frameworks like PyTorch
\citep{pytorch-docs} already provide pre-trained models with quantized
variants, and while PTQ is expected to degrade precision and performance,
empirical results show only minimal accuracy drops
\citep{nagel2021white,askarihemmat2024qgen}. This calls for an
investigation into how post-training quantization affects overall model
behaviour and feature selection; specifically, how reduced precision
influences the model's ability to capture critical data patterns,
which, to the best of our knowledge, has hitherto remained unexplored.
Since such effects are not readily observable through accuracy alone,
interpretability methods provide a valuable means of comparing the
feature selection and decision-making processes of full-precision and
post-training quantized models.

Early studies on neural network quantization focused on statistical
models to characterize performance degradation with reduced precision.
\citet{xie1992analysis} proposed a theoretical framework and predicted
that model performance would decline with decreasing bit precision.
However, subsequent empirical studies
\citep{krishnamoorthi2018quantizing,gholami2022survey,jacob2018quantization}
on CNN quantization found that quantization can often preserve
near-full precision, indicating that such performance degradation is not
inevitable in practice. \citet{dundar1995effects} extended this line of
research by considering bounded weights and sigmoid neurons, revealing an
approximately exponential relationship between accuracy and weight
precision. However, both studies primarily examined quantization at the
network output level and relied on several simplifying assumptions.
Furthermore, these studies did not investigate layer or feature-level
sensitivity, leaving a limited understanding of how quantization affects
internal representations and feature attribution.

\citet{rogers2023evaluating} investigated the impact of quantization on
the interpretability of convolutional neural networks (CNNs) by
introducing a quantization-aware explainable artificial intelligence
(XAI) framework. This framework compared activation patterns and heatmap
interpretations based on class activation maps (CAMs) between quantized
and full-precision models. The authors employed a pre-trained quantized
model and EigenCAM \citep{bany2021eigen} trained on the ImageNet dataset
\citep{deng2009imagenet}. This study focused primarily on visual
analysis, demonstrating that the class activation maps
\citep{zhou2016learning} of the quantized model produced more compact
bounding boxes around objects compared to the more dispersed regions
identified by the full-precision model. Similarly,
\citet{kerkouri2024quantization} examined the impact of neural network
quantization on the perceptual field of visual models, which is critical
for interpretability in resource-constrained deployments. This study
focused exclusively on quantization-aware training (QAT). By comparing
CAMs and visually salient object maps across three quantization levels,
it analyzed how quantization influences the spatial recognition
capability of various CNN architectures. The
results revealed subtle yet significant variations in CAMs, with
different architectures exhibiting varying degrees of sensitivity. For
example, SqueezeNet-1.0 maintained strong robustness, whereas
EfficientNet-B0 was highly sensitive to quantization, highlighting the
need to consider quantization effects when deploying models efficiently
while preserving interpretability.

While earlier studies \citep{xie1992analysis,dundar1995effects} made
important theoretical contributions, their reliance on simplified
statistical assumptions regarding activation distributions and noise
propagation limits their applicability to modern deep architectures. In
contrast, recent interpretability studies
\citep{rogers2023evaluating,kerkouri2024quantization} primarily rely on
CAM-based analysis, which focuses on a single layer within the neural
network and fails to capture the cumulative effects of quantization
across the entire network architecture. Consequently, these approaches do
not reveal how layer-wise variations interact to influence overall model
behaviour. Also, existing work has primarily focused on QAT,
whereas PTQ, a widely adopted and training-free approach, has received
comparatively little attention.

In this paper, we establish a comprehensive
evaluation framework for assessing whether or how static PTQ affects explainability / interpretability. In particular, we employ a CAM-based interpretability method (Grad-CAM
\citep{selvaraju2017grad}), which generates layer-specific class
activation maps for selected convolutional layers, and further complement
this analysis with the perturbation-based technique LIME (Local
Interpretable Model-agnostic Explanations) \citep{ribeiro2016should} to
provide a more comprehensive evaluation of how PTQ affects overall model
behaviour.\\

The key contributions of this study are summarized as follows:

\begin{itemize}
\item \textbf{Systematic PTQ interpretability evaluation}: We establish a quantitative evaluation framework for assessing explanation / interpretability  preservation under PTQ across multiple CNN architectures and precision levels

\item \textbf{Dual interpretability framework}: We employ LIME for
      input-level feature attribution and Grad-CAM for spatial attention
      analysis, thus capturing local decision boundaries and global
      activation patterns under quantization conditions.
\item \textbf{Practical deployment insights}: We systematically evaluate
      the trade-offs among different architectures under quantized
      settings, helping professionals make informed deployment decisions
      while understanding potential behavioural changes.
\end{itemize}

\section{Methodology}

\subsection{Datasets}

We used two binary classification datasets that present intricate visual
recognition tasks for the experimental evaluation. The first dataset is a
subset of the Food-101 dataset \citep{bossard14} comprising 1,250 images
of two classes, namely, pizza and steak. The second dataset we have used
is a subset of Poribohon-BD \citep{tabassum2020poribohon}, containing
1,000 images of two classes, bike and bicycle. Both datasets represent a
challenging binary classification task due to high intra-class
variability and require fine-grained feature recognition. Importantly,
Poribohon-BD represents a domain-adapted test case, compared to the
models' original training data; by further training the pre-trained
models on this Bangladesh-specific transportation dataset, we evaluate
quantization robustness in a domain-specific setting.

\subsection{Models}

To comprehensively evaluate the impact of quantization across diverse
architectural paradigms, we choose five widely-adopted CNN architectures:
VGG19 \citep{simonyan2014very}, ResNet18 \citep{he2016deep},
EfficientNet-B0 \citep{tan2019efficientnet}, MobileNetV2
\citep{sandler2018mobilenetv2}, and DenseNet161
\citep{huang2017densely}. These models represent fundamentally different
design principles, such as VGG's simple deep architecture, ResNet's
residual connections, EfficientNet's compound scaling, MobileNet's
depthwise separable convolutions, and DenseNet's dense connectivity
patterns. This selection enables analysis of how quantization effects
vary across architectures, computational complexity and parameter
distributions. We used pre-trained models from PyTorch, and then
fine-tuned them on the specific datasets.

\subsection{Quantization}

Quantization can be implemented in two ways, namely, Post-Training
Quantization (PTQ) and Quantization-Aware Training (QAT). PTQ is often
the preferred option when computational resources are limited and quick
deployment matters, such as on edge devices, embedded systems, or mobile
platforms. It compresses pre-trained models into low-precision formats
without requiring full retraining. As a result it significantly lowers
memory and compute costs while still delivering acceptable performance
\citep{Saha2025_EfficientQuant,Gordon2023_EPTQ,Lee2023_QHyViT,Kim2024_HyQ,Zhong2024_ERQ,Li2023_RepQViT}.
In PTQ a pretrained floating-point model is converted to lower precision
using a small calibration dataset (distinct from the training set) that
is used only to estimate the statistical distributions of weights and
activations; these statistics guide the choice of quantization parameters
(scales, zero-points) and observer settings.

\begin{figure}
  \centering
  \includegraphics[width=\linewidth]{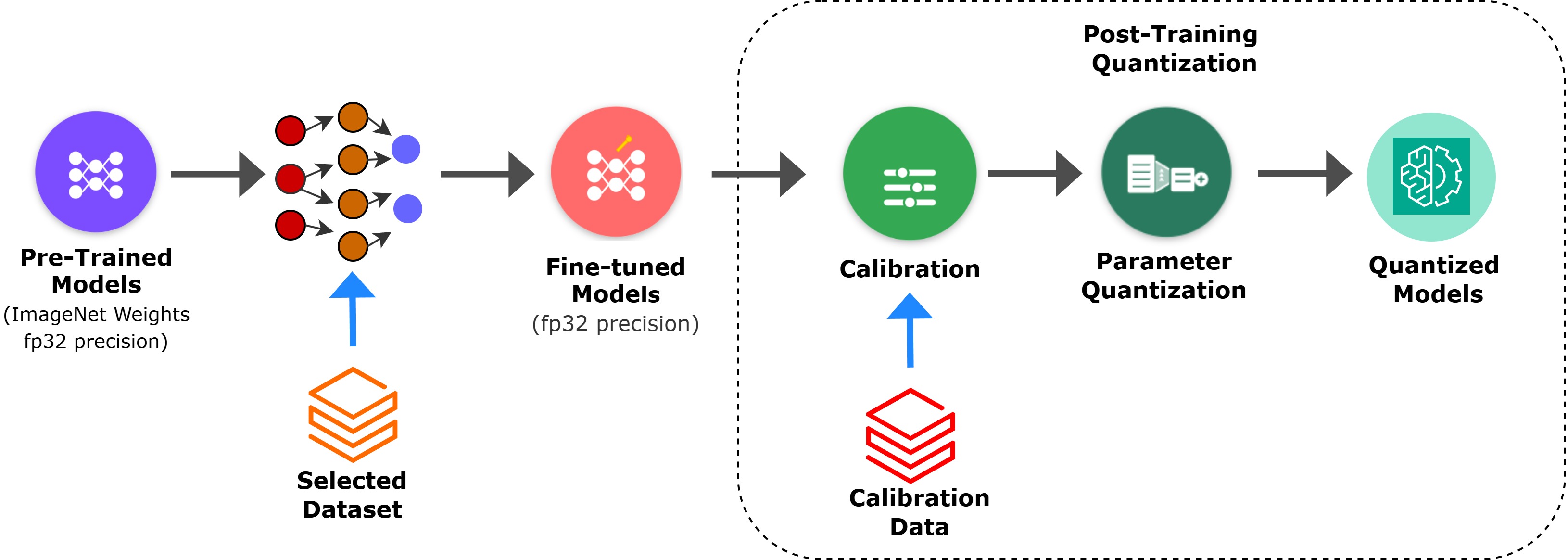}
  \caption{Post-training static quantization workflow. Pre-trained models
  (fp32) were fine-tuned on selected datasets and statically quantized to
  INT8/INT4. This reduces memory consumption and computation while
  maintaining accuracy, enabling efficient, retraining-free deployment.}
  \label{fig:quant1}
\end{figure}

PTQ can be done targeting only the model weights or both weights and
activations. We targetted weigt-only quantization and to achieve that we implemented GPFQ (Greedy Path Following Quantization) method \citep{zhang2023post} , which quantizes each neuron independently by
selecting quantized weights that minimize the difference between the
original and quantized layer outputs. This process is applied to all
selected models, resulting in INT8 and INT4 versions of each model.
Figure~\ref{fig:quant1} illustrates the post-training quantization
process employed in this study.

\subsection{Interpretability Techniques}

Interpretability methods aim to make neural networks more transparent by
explaining their internal mechanisms and decision-making processes. These
methods help identify which input features a model relies on and why it
makes certain predictions. CAM-based interpretability methods generate
class-specific activation maps that highlight the spatial regions most
influential to a model's decision. Perturbation-based methods, on the
other hand, work differently: they assess feature significance by
changing inputs and observing the resulting changes in model outputs. In
our experiment, we used both Grad-CAM \citep{selvaraju2017grad} for
gradient-based CAM analysis and LIME \citep{ribeiro2016should} for
perturbation-based attribution.

\paragraph{\textbf{Grad-CAM.}} Class Activation Maps (CAMs) were generated
using Grad-CAM, introduced by \citet{selvaraju2017grad}, to visualize and
compare the features that are most influential to the models'
predictions before and after quantization. Grad-CAM operates by computing
the gradient of the target class score with respect to the feature maps
of the final convolutional layer. These gradients are then globally
averaged to obtain weights that represent the importance of each feature
map for the target class. A weighted sum of the feature maps is computed
using these weights, followed by applying a ReLU activation to produce a
heatmap that highlights the image regions most relevant to the model's
decision, thereby offering insights into how quantization affects spatial
features.

\paragraph{\textbf{LIME.}} \citet{ribeiro2016should} introduced LIME,
which works by creating many slight perturbations of the input data and
obtaining predictions from the black-box model for each perturbed sample.
It then assigns higher weights to samples closer to the original instance
and fits a simple surrogate model locally around that instance. The
coefficients of this local model show which features contributed most to
the black-box model's prediction. In our experiments we have implemented
LIME to demonstrate the difference in features being selected by each of
the models with their quantized counterparts. We generated LIME explanation heatmaps for each image to visualize the image regions contributing most to the model's predictions (Figure \ref{fig:quant3}). We also used different metrics
to showcase the explanation of LIME in a mathematical approach.

\subsection{Evaluation Metrics}

\begingroup
\makeatletter
\@fleqnfalse
\makeatother

We employ a comprehensive evaluation framework to assess how quantization
affects model interpretability and decision-making processes. Our
analysis seeks to determine whether quantized models preserve the
attention patterns, activation structures, and feature representations of
their full-precision counterparts. In particular, we employ the following
metrics.

\subsubsection{Pearson Correlation Coefficient (PCC)}

For our heatmap vectors, the Pearson correlation \textit{(r)}
\citep{pearson1896vii} evaluates the linear correspondence between
activation patterns, capturing how proportional changes in one heatmap
correspond to changes in another. PCC is measured as follows:
\begin{equation}
r = \frac{\sum_{i=1}^{n} (x_i - \bar{x})(y_i - \bar{y})}
         {\sqrt{\sum_{i=1}^{n} (x_i - \bar{x})^2}
          \sqrt{\sum_{i=1}^{n} (y_i - \bar{y})^2}}
\end{equation}
where $n$ is the total number of observations, $x_i$ and $y_i$ represent
the $i$-th observations of variables $x$ and $y$, respectively, and
$\bar{x}$ and $\bar{y}$ denote the mean values of variables $x$ and $y$.
The summation $\sum_{i=1}^{n}$ is taken over all observations from $i=1$
to $n$.

\subsubsection{Structural Similarity (SSIM)}

The Structural Similarity Index (SSIM) \citep{wang2004image} quantifies
perceptual similarity between two heatmaps or feature vectors by jointly
evaluating luminance, contrast and structural information. Unlike
pixel-wise metrics, SSIM captures local patterns and structural
relationships consistent with the human visual system's sensitivity to
distortions. Higher SSIM values (closer to 1) indicate stronger
structural and perceptual similarity between the compared
representations. For two signals $x$ and $y$, SSIM is defined as:
\begin{equation}
\text{SSIM}(x, y) = \frac{(2\mu_x\mu_y + C_1)(2\sigma_{xy} + C_2)}
         {(\mu_x^2 + \mu_y^2 + C_1)(\sigma_x^2 + \sigma_y^2 + C_2)}
\end{equation}
where $\mu_x$ and $\mu_y$ are the mean intensities of $x$ and $y$,
$\sigma_x^2$ and $\sigma_y^2$ are their variances, $\sigma_{xy}$ is the
covariance between $x$ and $y$, and $C_1$ and $C_2$ are small constants
added for numerical stability.

\subsubsection{Top-$K$ IoU}

We adapt the standard Intersection-over-Union~\citep{rezatofighi2019giou}
to measure the spatial agreement between the explanation maps of the
full-precision and quantized models. For each image, we binarize both
heatmaps by retaining their top $K\%$ most important pixels, yielding
masks $M_{\text{fp32}}$ and $M_q$, and compute their
Intersection-over-Union:
\begin{equation}
\text{Top-}K\%\ \text{IoU} =
\frac{|M_{\text{fp32}} \cap M_q|}{|M_{\text{fp32}} \cup M_q|}
\end{equation}
Values near 1 indicate the two models emphasize the same regions, while
values near 0 indicate quantization has relocated the salient region. We
use $K = 20$ and report the mean across all images per cell.

\endgroup

\begin{fixedfigure}
  \centering
  \includegraphics[width=\linewidth]{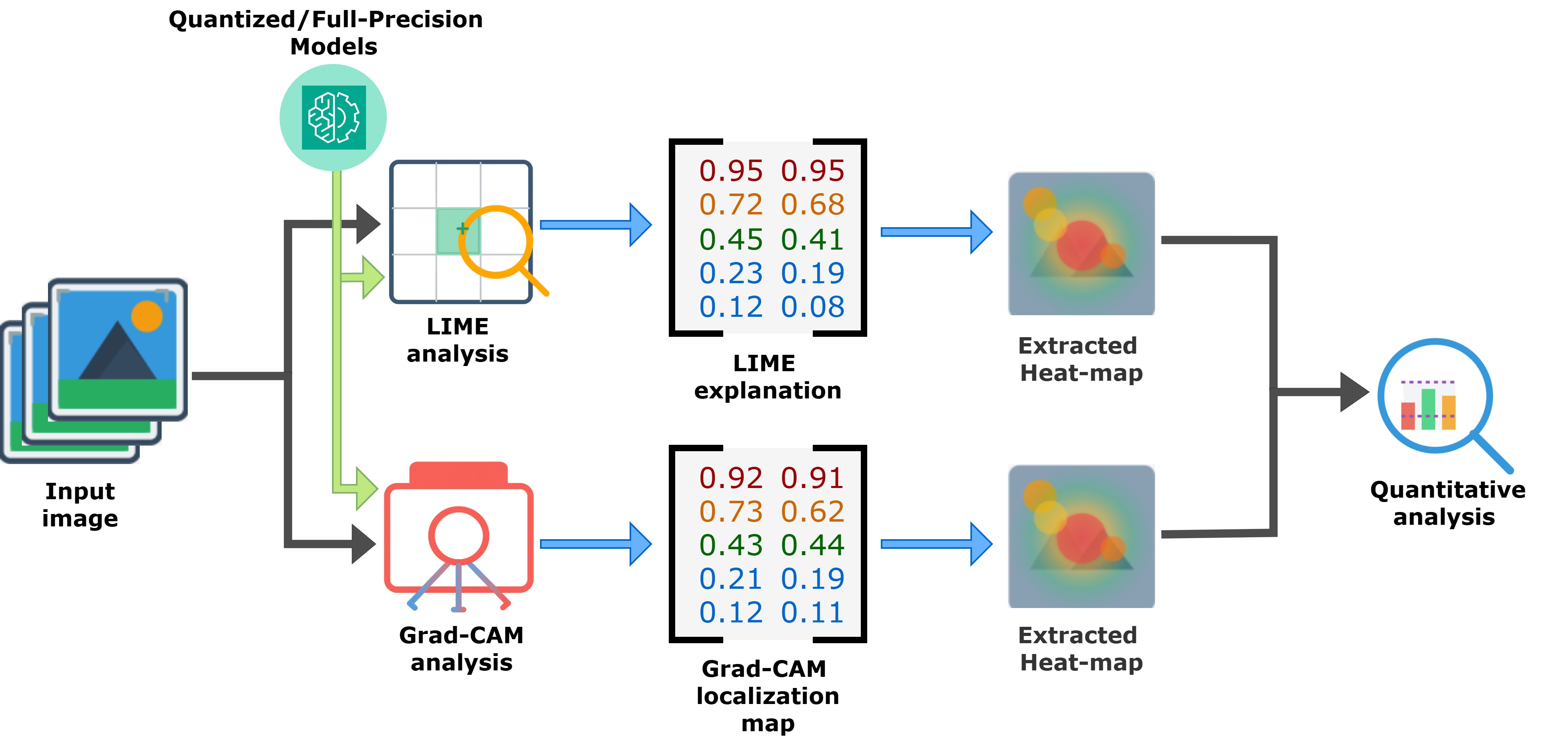}
  \caption{Interpretability analysis pipeline for quantized and full-precision models. Input
  images are analyzed by both LIME
  and Grad-CAM. Heatmaps from both processess are compared using different
  evaluation metrics to assess how quantization affects attention and
  decision-making.}
  \label{fig:quant2}
\end{fixedfigure}

\section{Results and Discussion}

Our empirical analysis reveals clear patterns
in how different CNN architectures retain interpretable features under
Post-Training Quantization (PTQ). As alluded above, to quantify the
alignment between full-precision and quantized model explanations across
both LIME and Grad-CAM frameworks, we examined three complementary
metrics: Pearson Correlation Coefficient (PCC), Structural Similarity
Index (SSIM) and Top-$K$ IoU. Tables~\ref{tab:sim_scores} and
\ref{tab:primary_stat} present a comprehensive comparison of the selected
metrics for both Grad-CAM and LIME across our two datasets.

\begin{table*}[h]
\centering
\caption{Explanation similarity to fp32, pooled across FOOD-101 and
Poribohon-BD. Cells report mean $\pm$ std of per-image similarity scores
(PCC, SSIM, and top-20\% IoU).}
\label{tab:sim_scores}
\resizebox{\textwidth}{!}{%
\begin{tabular}{llccccc}
\toprule
Metric & Comparison & MobileNetV2 & VGG19 & ResNet18 & EfficientNet-B0 & DenseNet161 \\
\midrule
\multicolumn{7}{l}{\textbf{Grad-CAM}} \\
\midrule
\multirow{3}{*}{PCC $\uparrow$}
& fp32 vs INT8 & 0.811 $\pm$ 0.229 & 0.982 $\pm$ 0.127 & 0.878 $\pm$ 0.159 & 0.670 $\pm$ 0.291 & 0.977 $\pm$ 0.147 \\
& fp32 vs INT4 & 0.748 $\pm$ 0.273 & 0.958 $\pm$ 0.127 & 0.715 $\pm$ 0.233 & 0.593 $\pm$ 0.318 & 0.975 $\pm$ 0.156 \\
& INT8 vs INT4 & 0.768 $\pm$ 0.273 & 0.958 $\pm$ 0.115 & 0.700 $\pm$ 0.229 & 0.653 $\pm$ 0.305 & 0.995 $\pm$ 0.052 \\
\midrule
\multirow{3}{*}{SSIM $\uparrow$}
& fp32 vs INT8 & 0.796 $\pm$ 0.226 & 0.980 $\pm$ 0.123 & 0.866 $\pm$ 0.160 & 0.639 $\pm$ 0.282 & 0.975 $\pm$ 0.143 \\
& fp32 vs INT4 & 0.733 $\pm$ 0.267 & 0.951 $\pm$ 0.121 & 0.694 $\pm$ 0.229 & 0.559 $\pm$ 0.306 & 0.974 $\pm$ 0.151 \\
& INT8 vs INT4 & 0.754 $\pm$ 0.268 & 0.952 $\pm$ 0.110 & 0.680 $\pm$ 0.226 & 0.624 $\pm$ 0.294 & 0.995 $\pm$ 0.048 \\
\midrule
\multirow{3}{*}{IoU $\uparrow$}
& fp32 vs INT8 & 0.571 $\pm$ 0.222 & 0.882 $\pm$ 0.150 & 0.641 $\pm$ 0.194 & 0.473 $\pm$ 0.211 & 0.881 $\pm$ 0.152 \\
& fp32 vs INT4 & 0.506 $\pm$ 0.218 & 0.778 $\pm$ 0.161 & 0.482 $\pm$ 0.201 & 0.426 $\pm$ 0.226 & 0.876 $\pm$ 0.155 \\
& INT8 vs INT4 & 0.527 $\pm$ 0.229 & 0.778 $\pm$ 0.163 & 0.470 $\pm$ 0.200 & 0.474 $\pm$ 0.218 & 0.931 $\pm$ 0.107 \\
\midrule
\multicolumn{7}{l}{\textbf{LIME}} \\
\midrule
\multirow{3}{*}{PCC $\uparrow$}
& fp32 vs INT8 & 0.384 $\pm$ 0.104 & 0.963 $\pm$ 0.019 & 0.825 $\pm$ 0.048 & 0.228 $\pm$ 0.128 & 0.987 $\pm$ 0.008 \\
& fp32 vs INT4 & 0.284 $\pm$ 0.113 & 0.831 $\pm$ 0.077 & 0.556 $\pm$ 0.106 & 0.109 $\pm$ 0.112 & 0.984 $\pm$ 0.010 \\
& INT8 vs INT4 & 0.338 $\pm$ 0.119 & 0.821 $\pm$ 0.083 & 0.562 $\pm$ 0.105 & 0.162 $\pm$ 0.139 & 0.994 $\pm$ 0.004 \\
\midrule
\multirow{3}{*}{SSIM $\uparrow$}
& fp32 vs INT8 & 0.333 $\pm$ 0.092 & 0.942 $\pm$ 0.026 & 0.763 $\pm$ 0.057 & 0.195 $\pm$ 0.104 & 0.979 $\pm$ 0.014 \\
& fp32 vs INT4 & 0.242 $\pm$ 0.101 & 0.768 $\pm$ 0.083 & 0.485 $\pm$ 0.095 & 0.103 $\pm$ 0.092 & 0.974 $\pm$ 0.016 \\
& INT8 vs INT4 & 0.294 $\pm$ 0.106 & 0.757 $\pm$ 0.090 & 0.489 $\pm$ 0.095 & 0.143 $\pm$ 0.112 & 0.990 $\pm$ 0.007 \\
\midrule
\multirow{3}{*}{IoU $\uparrow$}
& fp32 vs INT8 & 0.168 $\pm$ 0.045 & 0.692 $\pm$ 0.084 & 0.432 $\pm$ 0.082 & 0.143 $\pm$ 0.045 & 0.799 $\pm$ 0.078 \\
& fp32 vs INT4 & 0.146 $\pm$ 0.045 & 0.439 $\pm$ 0.100 & 0.235 $\pm$ 0.069 & 0.128 $\pm$ 0.042 & 0.780 $\pm$ 0.085 \\
& INT8 vs INT4 & 0.165 $\pm$ 0.058 & 0.431 $\pm$ 0.097 & 0.244 $\pm$ 0.070 & 0.132 $\pm$ 0.044 & 0.859 $\pm$ 0.068 \\
\bottomrule
\end{tabular}}
\end{table*}

\subsection{Quantitative Analysis}

\paragraph{\textbf{8-bit Quantization.}}Figure~\ref{fig:quant2} illustrates the quantitative analysis pipeline
implemented in this study. Quantization from fp32 to INT8 produces only mild interpretability changes for most of the selected architectures, although the degree varies. Tables~\ref{tab:sim_scores} report the average results obtained from both datasets. DenseNet-161 holds up best across both explainers. Its LIME attributions remain almost perfectly aligned with the full-precision model, averaging a PCC of 0.987, SSIM of 0.979, and top-20\% IoU of 0.799. Its grad-CAM scores are equally stable at PCC 0.977 and SSIM 0.975, preserving nearly 98\% of its full-precision correlation and structure. The dense connectivity and feature reuse of DenseNet-161 makes it less vulnerable to quantization, keeping both local decision boundaries and spatial attention intact. VGG-19 follows closely, holding Grad-CAM PCC of 0.982 and SSIM of 0.980, with LIME PCC of 0.963 and SSIM of 0.942. Its straightforward sequential convolutions make it quite resilient toward precision changes. With Grad-CAM PCC of 0.878 (SSIM 0.866) and LIME PCC of 0.825 (SSIM 0.763) ResNet-18 is also stable. Its residual connections preserve the overall structure of feature importance despite a reduction in localization accuracy (LIME top-20\% IoU of 0.432). MobileNet-V2 reveals a sharper divergence between the two explainers. Its Grad-CAM stays moderate (PCC 0.811, SSIM 0.796), while LIME drops to a PCC of 0.384, showing only about 39\% correspondence with fp32 with SSIM near 0.333. Built around depthwise separable convolutions, its global attention patterns survive while local attributions distort substantially. EfficientNet-B0 is the most fragile architecture even at INT8. Although its Grad-CAM correlation remains moderate (PCC 0.670), its LIME attributions degrade severely, with PCC collapsing to 0.228 and top-20\% IoU dropping near 0.143, its correspondence with full precision is under 23\%. The contrast between relatively preserved spatial attention and heavily degraded feature attribution suggests that quantization introduces inconsistencies within the model's internal representations. These observations imply that stable accuracy under quantization does not always ensure stable reasoning, which might be a concern for interpretability sensitive deployments.

\paragraph{\textbf{4-bit Quantization.}} Reducing to INT4 introduces more pronounced degradation, though again the extent is architecture dependent. DenseNet-161 remains the standout, with virtually no loss compared to INT8, with LIME PCC of 0.984 and SSIM of 0.974, also Grad-CAM PCC of 0.975 and SSIM of 0.974. We notice a negligible drop of about 0.003-0.005, a cost for lower bit-width, again making it a dependable choice for environments where interpretability is the concern. VGG-19 stays strong on correlation and structure (Grad-CAM PCC 0.958, LIME PCC 0.831), but its localization weakens more visibly. LIME top-20\% IoU falls from 0.692 at INT8 to 0.440, roughly a 36\% relative drop. Moreover, ResNet-18 shows a steeper decline. Its LIME PCC drops from 0.825 to 0.556 and SSIM to 0.485, indicating that residual connections offer only partial protection to input-level reasoning, once precision is this aggressive. MobileNet-V2 degrades further on the input-level side. Its LIME PCC falls to 0.283 with top-20\% IoU near 0.146, that is roughly 28\% correspondence, but Grad-CAM holds comparatively steady (PCC 0.748). For ranking-based tasks its spatial focus may still be usable, but absolute attribution values are no longer reliable. EfficientNet-B0 experiences the most severe degradation among all evaluated architectures. Its Grad-CAM correlation slips to 0.593, but its LIME attributions effectively collapse to a PCC of 0.109 and SSIM of 0.103, barely 11\% correspondence with fp32. At this precision the model's input-level reasoning bears almost no linear relationship to its full-precision counterpart. Accuracy metrics may still look acceptable, but the decision process becomes substantially less interpretable, which is a meaningful risk in any safety-critical context.

\begin{table*}[t]
\centering
\caption{Statistical comparison of explanation similarity preservation
between INT8 and INT4 GPFQ quantization, pooled across FOOD-101 and
Poribohon-BD. For each (method, architecture, metric) cell, we compute
the per-image paired difference
$\mathrm{diff}=\mathrm{sim}(\mathrm{fp32},\mathrm{INT8})-\mathrm{sim}(\mathrm{fp32},\mathrm{INT4})$.
Normality of the paired differences is assessed using the
Shapiro--Wilk test, followed by a paired $t$-test when normally
distributed or a Wilcoxon signed-rank test otherwise. Effect size is
reported as paired Cohen's $d=\overline{\mathrm{diff}}/s_{\mathrm{diff}}$.
Positive $d$ indicates that INT8 explanations are more similar to fp32
explanations than INT4 explanations, corresponding to better explanation
preservation under INT8 quantization.}
\label{tab:primary_stat}
\footnotesize
\begin{tabular}{@{}lccccc@{}}
\toprule
Metric & MobileNetV2 & VGG19 & ResNet18 & EfficientNet-B0 & DenseNet161 \\
\midrule
\multicolumn{6}{l}{\textbf{Grad-CAM}} \\
\midrule
PCC $\uparrow$  & $d=0.26$ & $d=0.21$ & $d=0.81$ & $d=0.29$ & $d=0.03$ \\
SSIM $\uparrow$ & $d=0.27$ & $d=0.27$ & $d=0.88$ & $d=0.31$ & $d=0.03$ \\
IoU $\uparrow$  & $d=0.30$ & $d=0.63$ & $d=0.80$ & $d=0.24$ & $d=0.06$ \\
\midrule
\multicolumn{6}{l}{\textbf{LIME}} \\
\midrule
PCC $\uparrow$  & $d=0.82$ & $d=1.91$ & $d=3.12$ & $d=0.88$ & $d=0.45$ \\
SSIM $\uparrow$ & $d=0.86$ & $d=2.36$ & $d=3.54$ & $d=0.79$ & $d=0.44$ \\
IoU $\uparrow$  & $d=0.37$ & $d=2.52$ & $d=2.54$ & $d=0.29$ & $d=0.36$ \\
\bottomrule
\end{tabular}
\end{table*}

To confirm whether INT8 preserves fp32 explanations better than INT4 at 
a statistically meaningful level, we computed the paired difference 
$\mathrm{sim}(\mathrm{fp32}, \mathrm{INT8}) - \mathrm{sim}(\mathrm{fp32}, 
\mathrm{INT4})$ per image, tested normality with Shapiro--Wilk. For normally distributed samples, we applied a paired t-test; otherwise, we employed the non-parametric Wilcoxon signed-rank test. Effect sizes are reported as paired Cohen's d (Table~\ref{tab:primary_stat}). Across all 
five architectures and both explainers, the effect is positive and 
statistically significant at $p < 0.001$, confirming that INT8 stays closer 
to full-precision explanations than INT4 does, even when accuracy differences 
between the two are negligible. The magnitude is strongly 
architecture-dependent. DenseNet161 shows the smallest overall effect: its 
LIME explanations are nearly unchanged by the INT8-to-INT4 step ($d = 0.44$--$0.45$), 
and its Grad-CAM effects remain the lowest of any architecture on PCC ($d = 0.03$), 
making it the most bit-width-stable model across both explainers. ResNet18 
sits at the opposite end, with LIME effect sizes of $d = 3.12$ on PCC and 
$d = 3.54$ on SSIM, meaning INT8 is substantially closer to fp32 than INT4 
on every paired test image. VGG19 follows at $d = 1.91$ on LIME PCC, while 
MobileNetV2 and EfficientNet-B0 fall in an intermediate range ($d \approx 
0.8$ on LIME PCC). Grad-CAM effects follow the same architecture ordering 
but are uniformly smaller, peaking at $d = 0.88$ for ResNet18, confirming 
that gradient-based attributions are more stable to the INT8-to-INT4 
transition than perturbation-based ones.

\begin{figure}
  \centering
  \includegraphics[width=\textwidth]{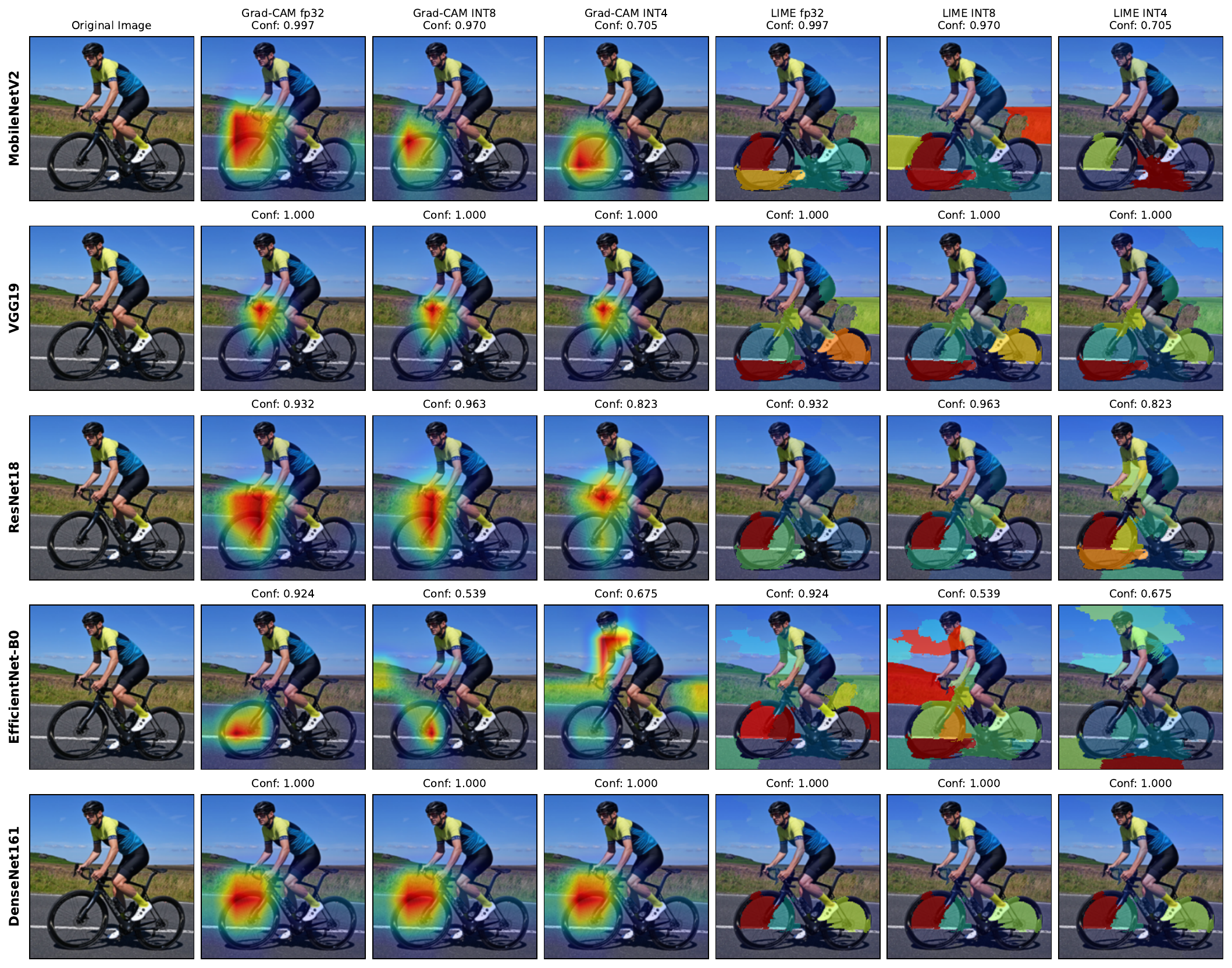}
  \caption{Comparison of Grad-CAM and LIME explanation heatmaps for
  multiple CNN architectures under different precision levels (float32,
  INT8, INT4). Rows represent models, and columns correspond to precision
  levels with associated confidence scores.}
  \label{fig:quant3}
\end{figure}

\subsection{Qualitative Analysis}

Figure~\ref{fig:quant3} shows the Grad-CAM and LIME heatmaps of VGG19, ResNet18, EfficientNet-B0, DenseNet-161, and MobileNetV2 at float32, INT8, and INT4 precisions. The image is taken from the Poribohon-BD dataset. The goal is to visually assess whether quantization shifts where a model focuses on when making its prediction.

VGG19 and DenseNet-161 are the most stable models among the chosen architectures. Across all three precision, the confidence score remains at 1.00, and the heatmap consistently focuses on the bicycle wheel and lower body region. The depth of these architectures does not appear to introduce any significant spatial drift under quantization. ResNet18 shows slight spread at INT8 but still focuses on the rider and bicycle frame. The confidence score remains between 0.823 and 0.963, and the overall coverage remains stable; therefore, this shift, while perceptible, does not affect the prediction results.

EfficientNet-B0 exhibits different patterns. The float32 model focuses tightly on the bicycle frame with a confidence level of 0.924. At INT8 precision, this focus expands, and the confidence level drops to 0.539. Different precision levels focus on entirely different features, consistent with the quantitative analysis that EfficientNet-B0's compound scaling makes its intermediate activations particularly sensitive to precision reduction. MobileNetV2 remains largely consistent between float32 and INT8. Activations expand at INT4, but the core region of interest does not change significantly. Even with changes in internal attribution strength, the spatial structure remains unchanged.

Overall, the visual evidence reinforces the paper's quantitative findings: most models tolerate quantization reasonably well at the spatial attention level, but EfficientNet-B0 stands out as a clear outlier where int8 quantization fundamentally disrupts the model's internal reasoning, even when accuracy metrics might suggest otherwise.

\subsection{Comparative Analysis}

\paragraph{\textbf{Faithfulness analysis.}}
We define faithfulness as
the extent to which an explanation identifies the regions that truly
drive the model's predictions. Table~\ref{tab:faith_merged} reports the Area Under the Curve (AUC) values for two metrics: \textbf{deletion}, which measures how classification confidence falls as the regions the explanation ranks most important are progressively removed from the image, and \textbf{insertion}, which measures how confidence rises as those same regions are added back. Together, these values test whether the explanation identifies regions that actually drive the model's predictions, rather than regions that are merely associated with them. In the deletion and insertion framework,
faithful explanations produce lower deletion AUC values, showing a strong
drop in confidence when important regions are removed, while higher
insertion AUC values indicate a fast recovery of confidence when these
regions are kept.

\begin{table*}[!htbp]
\centering
\caption{Faithfulness (deletion / insertion confidence-trajectory AUC),
pooled across FOOD-101 and Poribohon-BD images. Pixels are ranked in
descending order of explanation importance and perturbed in
$16\times16$ blocks over 50 steps. Deletion replaces pixels with a black
baseline, while insertion reveals them on a $\sigma=15$ Gaussian-blurred
baseline. We record the model's softmax confidence for the fp32-predicted
class along each curve and integrate trapezoidally. Lower deletion AUC
and higher insertion AUC indicate more faithful explanations. Cells
report mean $\pm$ std of the per-image AUC distribution pooled across
both datasets.}
\label{tab:faith_merged}
\resizebox{\textwidth}{!}{%
\begin{tabular}{llccccc}
\toprule
Metric & Precision & MobileNetV2 & VGG19 & ResNet18 & EfficientNet-B0 & DenseNet161 \\
\midrule
\multicolumn{7}{l}{\textbf{Grad-CAM}} \\
\midrule
\multirow{3}{*}{Deletion AUC $\downarrow$}
& fp32 & 0.611 $\pm$ 0.284 & 0.607 $\pm$ 0.201 & 0.628 $\pm$ 0.168 & 0.591 $\pm$ 0.162 & 0.596 $\pm$ 0.284 \\
& INT8 & 0.591 $\pm$ 0.277 & 0.605 $\pm$ 0.187 & 0.617 $\pm$ 0.169 & 0.565 $\pm$ 0.141 & 0.595 $\pm$ 0.286 \\
& INT4 & 0.611 $\pm$ 0.250 & 0.605 $\pm$ 0.187 & 0.611 $\pm$ 0.159 & 0.552 $\pm$ 0.144 & 0.598 $\pm$ 0.287 \\
\midrule
\multirow{3}{*}{Insertion AUC $\uparrow$}
& fp32 & 0.923 $\pm$ 0.080 & 0.948 $\pm$ 0.062 & 0.891 $\pm$ 0.103 & 0.835 $\pm$ 0.107 & 0.963 $\pm$ 0.044 \\
& INT8 & 0.923 $\pm$ 0.073 & 0.946 $\pm$ 0.068 & 0.883 $\pm$ 0.105 & 0.874 $\pm$ 0.095 & 0.962 $\pm$ 0.045 \\
& INT4 & 0.904 $\pm$ 0.112 & 0.942 $\pm$ 0.073 & 0.865 $\pm$ 0.112 & 0.843 $\pm$ 0.120 & 0.962 $\pm$ 0.045 \\
\midrule
\multirow{3}{*}{Gap (Ins$-$Del) $\uparrow$}
& fp32 & 0.312 $\pm$ 0.297 & 0.341 $\pm$ 0.193 & 0.263 $\pm$ 0.186 & 0.243 $\pm$ 0.182 & 0.368 $\pm$ 0.279 \\
& INT8 & 0.332 $\pm$ 0.262 & 0.341 $\pm$ 0.181 & 0.266 $\pm$ 0.192 & 0.309 $\pm$ 0.125 & 0.368 $\pm$ 0.283 \\
& INT4 & 0.293 $\pm$ 0.270 & 0.337 $\pm$ 0.177 & 0.254 $\pm$ 0.203 & 0.291 $\pm$ 0.147 & 0.364 $\pm$ 0.287 \\
\midrule
\multicolumn{7}{l}{\textbf{LIME}} \\
\midrule
\multirow{3}{*}{Deletion AUC $\downarrow$}
& fp32 & 0.810 $\pm$ 0.144 & 0.438 $\pm$ 0.151 & 0.714 $\pm$ 0.189 & 0.645 $\pm$ 0.119 & 0.830 $\pm$ 0.199 \\
& INT8 & 0.741 $\pm$ 0.151 & 0.460 $\pm$ 0.154 & 0.762 $\pm$ 0.160 & 0.651 $\pm$ 0.131 & 0.834 $\pm$ 0.200 \\
& INT4 & 0.733 $\pm$ 0.190 & 0.514 $\pm$ 0.159 & 0.713 $\pm$ 0.130 & 0.562 $\pm$ 0.100 & 0.836 $\pm$ 0.200 \\
\midrule
\multirow{3}{*}{Insertion AUC $\uparrow$}
& fp32 & 0.866 $\pm$ 0.116 & 0.863 $\pm$ 0.127 & 0.841 $\pm$ 0.104 & 0.800 $\pm$ 0.098 & 0.940 $\pm$ 0.082 \\
& INT8 & 0.894 $\pm$ 0.098 & 0.866 $\pm$ 0.126 & 0.849 $\pm$ 0.099 & 0.830 $\pm$ 0.122 & 0.931 $\pm$ 0.089 \\
& INT4 & 0.858 $\pm$ 0.126 & 0.851 $\pm$ 0.140 & 0.799 $\pm$ 0.101 & 0.703 $\pm$ 0.134 & 0.928 $\pm$ 0.091 \\
\midrule
\multirow{3}{*}{Gap (Ins$-$Del) $\uparrow$}
& fp32 & 0.056 $\pm$ 0.127 & 0.425 $\pm$ 0.173 & 0.127 $\pm$ 0.193 & 0.156 $\pm$ 0.130 & 0.110 $\pm$ 0.192 \\
& INT8 & 0.153 $\pm$ 0.136 & 0.406 $\pm$ 0.171 & 0.086 $\pm$ 0.155 & 0.179 $\pm$ 0.123 & 0.097 $\pm$ 0.200 \\
& INT4 & 0.125 $\pm$ 0.161 & 0.336 $\pm$ 0.185 & 0.086 $\pm$ 0.139 & 0.141 $\pm$ 0.128 & 0.092 $\pm$ 0.203 \\
\bottomrule
\end{tabular}}
\end{table*}

Our main observation is that faithfulness is preserved even at low-bit
quantization. Across both interpreters and all five architectures, the
deletion and insertion AUC values for the quantized models remain close
to their fp32 counterparts. As a result, the insertion--deletion gap stays
almost unchanged from fp32 to INT4 (e.g., 0.368 for DenseNet161 with
Grad-CAM at fp32 versus 0.364 at INT4). Since confidence is always
measured with respect to the category predicted by the fp32 model, this
stability indicates that quantization does not noticeably change the
features responsible for the original predictions, even at INT4
precision. Overall, this analysis shows that the feature regions identified by Grad-CAM and LIME correspond to features that are important for the model's decision-making, as reflected by higher insertion scores and lower deletion scores.

\paragraph{\textbf{Architecture-specific insights and deployment
recommendations.}}
Among all the interpretability frameworks and precision levels tested, DenseNet161 is least affected by quantization among the evaluated models. Its dense connectivity produces a stable feature space that quantization noise does not significantly interfere with. At INT8 and INT4 precision, the PCC and SSIM values remain above 0.96, meaning that the feature representation and spatial attention mechanism remain effective even under high compression. For applications where interpretability is critical (e.g., healthcare, autonomous systems, and regulatory domains), DenseNet161 can be the most reliable choice, with the efficiency gains from quantization more than offsetting its higher baseline computational cost.

ResNet18 and VGG19 perform well at INT8 precision, with PCC values between 0.86 and 0.99 for Grad-CAM and between 0.81 and 0.97 for LIME. However, their performance differs at INT4 level. VGG19's performance remained stable (LIME PCC approximately 0.81-0.96), while ResNet18's performance dropped significantly (LIME PCC decreased to approximately 0.50–0.61). VGG's sequential convolutional structure is more tolerant of aggressive quantization than ResNet's residual connections. Both are suitable for general deployment scenarios where interpretability is useful but not absolutely necessary. VGG19 can tolerate some performance degradation at INT4; if attribution fidelity is critical, ResNet18 should remain at INT8.

MobileNetV2's results were mixed. Under Grad-CAM, its spatial consistency was acceptable (PCC approximately 0.81 at INT8, decreasing to 0.75 at INT4), but LIME attribution performance was weak from the start (PCC approximately 0.39 at INT8) and continued to decline to approximately 0.28 at INT4. Spatial focus remains relatively stable, but local attribution quality is inconsistent. If MobileNetV2 is deployed in a quantized environment, post-quantization interpretability validation is worthwhile, especially since calibrating feature attribution values at INT4 precision is crucial.

When interpretability is a deployment requirement, EfficientNet-B0 is the least suitable architecture. Grad-CAM explanation differs moderately (PCC approximately 0.67 at INT8 and approximately 0.60 at INT4), but LIME attribution performance drops sharply: PCC is 0.23 at INT8 accuracy and only 0.11 at INT4 accuracy. Its compound scaling strategy makes intermediate feature representations highly sensitive to precision degradation, and this phenomenon is already apparent at INT8 accuracy, not just INT4. Classification accuracy may still be competitive, but the model's decision evidence has almost no correlation with full-precision behavior. If EfficientNet-B0 must be used in an environment with extremely high interpretability requirements, full precision should be maintained.

\paragraph{\textbf{Misclassifications after quantization.}}
Following post-quantization, a marginal reduction in classification
accuracy was observed, consistent with typical quantization effects. To
investigate the nature of quantization-induced errors, we analyzed the
disagreement subset, i.e., images that were correctly classified by the
fp32 model but misclassified after quantization. For these images, we
compared Grad-CAM heatmaps generated by the full-precision and quantized
models to determine whether prediction changes occurred despite relying
on similar visual evidence.

Across both datasets and quantization levels, heatmap similarity on the
disagreement subset was generally low. At INT8, mean PCC values ranged
from $-0.626$ to $-0.051$ on FOOD-101 and from $-0.358$ to $-0.108$ on
Poribohon-BD. Similar behaviour was observed at INT4, where mean PCC
values ranged from $-0.593$ to $-0.185$ on FOOD-101 and from $-0.337$ to $-0.173$ on Poribohon-BD. Top-20\% IoU values were likewise low, with
per-architecture means generally below $0.12$ across both datasets and
bit-widths. While architecture-specific differences were visible, the
overall trend was consistent: disagreement images exhibited substantial
spatial reorientation of saliency between the fp32 and quantized models.

These findings indicate that quantization-induced misclassifications are
typically not cases where the model attends to the same image regions but
produces a different label. Instead, prediction failures are generally
accompanied by a shift in the visual evidence used by the model,
resulting in corresponding changes in the Grad-CAM explanations.
Moreover, the similar PCC and IoU distributions observed at INT8 and INT4
suggest that reducing precision from 8-bit to 4-bit does not
fundamentally alter the nature of these errors. Rather, when quantization
causes a prediction change, the quantized model tends to rely on
different image regions than its full-precision counterpart.

\paragraph{\textbf{8-bit versus 4-bit trade-offs.}}
The choice between INT8 and INT4 ultimately involves a trade-off between
preserving fp32-style explanations and maximizing deployment efficiency.
INT8 typically generates explanations closer to the original fp32 model's
explanation; therefore, if the goal is to preserve the inference process
learned during training, INT8 is a better choice. On the other hand, INT4
offers a higher compression ratio while maintaining comparable
classification accuracy and explanation fidelity. Its explanations may
differ more significantly from the corresponding fp32 explanations, but
they still have causal relevance to the predictions of the deployed
model. Therefore, the real difference between these two levels of
precision lies not in fidelity, but in whether the explanation should
reflect the inference process of the original fp32 model or that of the
actually deployed compressed model.

\section{Conclusion}

This study aims to answer a fundamental question: \textit{does the
internal inference process of a model change when it is quantized to
lower precision?} Our results demonstrate that the gap between accuracy
preservation and interpretability preservation is often greater than
commonly assumed. By evaluating multiple metrics using Grad-CAM and LIME
heatmaps, we find that while post-training quantization largely preserves
classification accuracy, it can substantially alter how the model
actually arrives at its decisions.

Our analysis shows that DenseNet-161 exhibits the strongest architectural
stability across both interpretability methods, even at INT4 precision,
making it the most reliable choice when interpretability is critical
during deployment. While all the other selected models show similar and
moderate degradation, EfficientNet-B0 is the most vulnerable. Even though
its Grad-CAM performance remains acceptable, with only moderate changes,
LIME attribution deteriorates completely at INT4 precision, a subtle
issue that cannot be identified by accuracy metrics alone.

Our findings provide valuable insights into the deployment of quantized
models in resource-constrained environments. Understanding how feature
representations and inference behaviour change after quantization can
help researchers and practitioners carefully evaluate the associated
trade-offs and make informed deployment decisions. Nevertheless, this
study has several limitations. We conducted experiments on only two
binary classification tasks using subsets of the Food-101 and
Poribohon-BD datasets. Furthermore, we evaluated five CNN architectures
while excluding transformer-based models, whose attention mechanisms may
respond differently to reduced precision. Extending this evaluation
framework to multi-class classification tasks, more diverse datasets,
transformer architectures, and task-specific calibration represents a
promising direction for future research toward bridging the gap between
efficient inference and trustworthy deployment.

\bibliographystyle{casmodel2names}
\bibliography{casrefs}

\end{document}